\documentclass[manuscript,screen,review, sigconf, anonymous]{IEEEtran}
\IEEEoverridecommandlockouts

\usepackage[utf8]{inputenc} % allow utf-8 input
\usepackage[T1]{fontenc}    % use 8-bit T1 fonts
\usepackage{hyperref}       % hyperlinks
\usepackage{url}            % simple URL typesetting
\usepackage{booktabs}       % professional-quality tables
\usepackage{amsfonts}       % blackboard math symbols
\usepackage{nicefrac}       % compact symbols for 1/2, etc.
\usepackage{microtype}      % microtypography
\usepackage{lipsum}
\usepackage{fancyhdr}       % header
\usepackage{graphicx}       % graphics
\graphicspath{{media/}}     % organize your images and other figures under media/ folder
\usepackage{subcaption}  
\usepackage{graphicx}
\usepackage{adjustbox}
\usepackage{array}
\usepackage{booktabs}
\usepackage{xcolor}
\usepackage{amsmath}
\usepackage{soul}
\def\BibTeX{{\rm B\kern-.05em{\sc i\kern-.025em b}\kern-.08em
    T\kern-.1667em\lower.7ex\hbox{E}\kern-.125emX}}
\begin{document}

%% Title
\title{On the Privacy-Preserving Properties of Spiking Neural Networks with Unique Surrogate Gradients and Quantization Levels}

\author{\IEEEauthorblockN{Ayana Moshruba, Shay Snyder, Hamed Poursiami, Maryam Parsa\footnote{Corresponding author}}\\
\IEEEauthorblockA{\textit{Electrical and Computer Engineering} \\
\textit{George Mason University}\\
Fairfax, USA \\
\{amoshrub, ssnyde9, hpoursia, mparsa\}@gmu.edu}
}
% \author{\IEEEauthorblockN{Anonymous Authors}}
\maketitle

\begin{abstract}

As machine learning models increasingly process sensitive data, understanding their vulnerability to privacy attacks is vital. Membership Inference Attack (MIA), which infer whether specific data points were used during training, is one such privacy risk. Previous work suggests that Spiking Neural Networks (SNNs), which rely on event-driven computation and discrete spike-based encoding, exhibit greater resilience to MIAs compared to Artificial Neural Networks (ANNs). This resilience is attributed to their non-differentiable activations and inherent stochasticity, which reduce the correlation between model responses and individual training samples. To further enhance privacy in SNNs, we explore two techniques: Quantization and Surrogate Gradients. Quantization, which reduces model precision to limit information leakage, has been shown to improve privacy resilience in ANNs. Since SNNs exhibit sparse and irregular activations, quantization may have an even stronger effect on disrupting the activation patterns exploited by MIAs. In this study, we compare the vulnerability of SNNs and ANNs to MIAs under \textit{weight} and \textit{activation} quantization across multiple datasets. We evaluate privacy vulnerability using the attack model’s Receiver Operating Characteristic (ROC) curve’s Area Under the Curve (AUC) metric, where lower values indicate stronger privacy protection, and assess model accuracy to quantify the privacy-accuracy trade-off. Our results show that quantization enhances privacy in both architectures with minimal performance degradation, but full-precision SNNs remain more resilient than even quantized ANNs.  Additionally, we examine the impact of surrogate gradients on privacy in SNNs. Among the five surrogate gradients evaluated, Spike Rate Escape provides the best privacy-accuracy trade-off, while Arctangent (aTan) increases vulnerability to MIAs. These findings reinforce SNNs’ inherent privacy advantages and demonstrate that both quantization and surrogate gradient selection can further influence privacy-accuracy trade-offs in SNNs.

\end{abstract}

\begin{IEEEkeywords}
quantization, surrogate gradients, spiking neural networks, membership inference attack, adversarial machine learning
\end{IEEEkeywords}

\section{Introduction}

\noindent
The widespread adoption of Machine Learning (ML) across domains such as healthcare~\cite{abouelmehdi2017big}, finance~\cite{tripathi2020financial}, and education~\cite{florea2020big} has raised concerns about privacy risks,  particularly from attacks exploiting model behaviors to infer sensitive information~\cite{liu2021machine, dwork2017exposed}. Artificial Neural Networks (ANNs) are widely used across various domains but remain vulnerable to privacy attacks, like Membership Inference Attacks (MIAs), where adversaries attempt to infer whether specific data points were used during training~\cite{golla2023security}. Prior study suggests that Spiking Neural Networks (SNNs) show lower vulnerability to MIAs compared to ANNs~\cite{Moshruba2024AreNA}, potentially due to their unique spike based computation, which introduces inherent privacy advantages~\cite{poursiami2024brainleaks}. Specifically, the non-differentiable and discontinuous nature of SNNs reduces the correlation between model responses and individual data points~\cite{meng2022training}, while their stochastic spike-based encoding increases representation diversity, reducing the risk of overfitting and making individual training samples less distinguishable~\cite{olin2021stochasticity}. 

Given these properties, we explore methods to further enhance the privacy of SNNs. First, we investigate quantization, which has been proven to improve privacy in ANNs by reducing precision and altering learned representations, potentially limiting information leakage~\cite{famili2023deep}. Since SNNs operate in an event-driven manner, where neurons fire only when necessary~\cite{schuman2022opportunities}, the combination of spiking sparsity and quantization may further disrupt activation patterns exploited by MIAs. Additionally, we hypothesize that the information loss and noise introduced by quantization~\cite{kang2024effect} could influence the model’s susceptibility to privacy attacks.

Second, we examine the role of surrogate gradients in shaping SNN privacy. These functions approximate gradients for non-differentiable spike events, enabling backpropagation-based training. Beyond facilitating learning, surrogate gradients introduce inherent variability into model outputs through spike timing distributions and gradient approximations. This stochasticity resembles the noise injection mechanisms of Differentially Private Stochastic Gradient Descent (DPSGD)~\cite{song2013stochastic}, suggesting a potential privacy-preserving effect. To explore this, we analyze five different surrogate gradient functions: Fast Sigmoid, Arctangent (aTan), Spike Rate Escape(STE), Triangular, and Straight Through Estimator \cite{eshraghian2021training} and evaluate their influence on privacy vulnerability and model accuracy.

In this study, we assess the effects of both \textit{weight} and \textit{activation} quantization on the privacy characteristics of SNNs and ANNs against MIAs, evaluating models across convolutional and fully connected architectures. Our experiments span five datasets: CIFAR-10~\cite{cifar10}, MNIST~\cite{lecun2010mnist}, FMNIST~\cite{xiao2017fashion}, Iris~\cite{omelina2021survey}, and Breast Cancer~\cite{misc_breast_cancer_14}. Privacy vulnerability is measured using the attack model’s Receiver Operating Characteristic (ROC) Area Under the Curve (AUC), where lower values indicate stronger privacy protection. We also assess model accuracy to quantify the trade-off between privacy and performance under different quantization and surrogate gradient settings.

Based on our experimental analysis, we summarize the key findings regarding the impact of quantization and surrogate gradients on privacy vulnerability and model performance below.

\begin{itemize}
    \item \textbf{Quantization Analysis:} Both weight and activaction quantization reduces MIA vulnerability compared to full precision in ANNs and SNNs, while introducing  minor degradation in model performance. However, extreme quantization to 2 bits imposes unnecessary performance degradation without substantial privacy benefits, making moderate quantization (4-bit and 8-bit) a more practical choice.

    \item \textbf{Privacy Protection:} While quantization enhances privacy protection in both architectures, full precision SNNs demonstrate inherently superior privacy compared to quantized ANNs across all datasets, highlighting their fundamental privacy advantage without compromising model performance.
    \item \textbf{Surrogate Gradient Evaluation:} Spike rate escape demonstrates superior privacy protection while maintaining high model accuracy, whereas  aTan and STE exhibit higher MIA vulnerability.
\end{itemize}

\section{Related Work}

% \hl{we need to mention gap in related work. You have referred to all these papers and then suddendly end the section. then what? what is the gap in these works that you are addressing in this paper}

% \hl{MP: Maybe you have written and I'm just too tired now, but it is not clear in the related work the difference between weight and activation quantiziation in ANN (or SNN) community. When you say quantized ANN which one are you referring to? How do you compare performance of weight-quantizied and activation-quantized ANN/SNNs?}

\noindent
Privacy attacks targeting sensitive data have raised concerns about information exposure, driving research into vulnerabilities like MIAs. Li et al.~\cite{liu2021machine} provide an overview of privacy attacks, including MIAs, which were first introduced in genomic studies by Homer et al.~\cite{homer2008resolving}. Shokri et al.~\cite{shokri} later formalized MIAs by developing the shadow model framework, where labeled datasets are generated to train an attack model for inferring data membership. Refinements by Salem et al.~\cite{salem2018ml} showed that MIAs could rely on single shadow models and confidence scores, while Nasr et al.~\cite{nasr2019comprehensive} demonstrated that white-box access enhances attack precision. While most privacy research has focused on ANNs, studying MIAs in SNNs is valuable due to their potential resilience, which comes from their sparse activations, where neurons fire only when necessary and the randomness in spike timing, both of which make it harder for an attacker to infer data membership. Notably, prior work highlights that SNNs incur less performance degradation when employing DPSGD compared to ANNs~\cite{Moshruba2024AreNA}. Han et al.~\cite{han2023towards} and Safronov et al.~\cite{kim2022privatesnn} introduced privacy-preserving techniques like federated learning and differential privacy for SNNs.

While prior research has explored privacy vulnerabilities in SNNs and introduced privacy-preserving techniques, further investigation is needed to understand how architectural modifications can enhance their resilience against MIAs. One such modification is quantization which has become an effective approach for reducing model size and computational demands, particularly in resource-constrained settings~\cite{yang2019quantization}. By reducing weight and activation precision, quantization introduces noise, which  protect privacy by making adversarial inference more challenging~\cite{kang2024effect}. Studies on ANNs using DoReFa-Net~\cite{kowalski2022towards, famili2023deep, wei2024q} apply both weight and activation quantization, demonstrating reduced MIA success rates due to added noise and lower overfitting. Applying quantization to SNNs has proven valuable for improving energy efficiency. Yin et al.~\cite{yin2024mint} demonstrated that weight and membrane potential quantization in SNNs reduces memory use by 93.8\% and computation energy by 90\% with minimal accuracy loss. Schaefer et al.~\cite{schaefer2020quantizing, schaefer2023hardware} showed that ternary quantization optimizes hardware efficiency by reducing energy and memory costs. Frameworks like Q-SpiNN~\cite{putra2021q} and QFFS~\cite{li2022quantization} adapt quantization techniques to SNN-specific dynamics, addressing challenges like synaptic weight and membrane potential quantization for low-power neuromorphic systems. Recent findings suggest that quantization noise in SNNs may reduce information leakage in MIAs, aligning with observations in ANNs~\cite{kang2024effect, famili2023deep}. 

On the other hand, surrogate gradients enable effective training in SNNs by approximating gradients for non-differentiable spike events,  but their impact on privacy has received little attention. PrivateSNN, introduced by Kim et al., proposes a privacy-preserving approach for converting ANNs to SNNs by incorporating spike-based learning rules to mitigate privacy risks~\cite{kim2022privatesnn}. Similarly, DPSNN, developed by Wang et al., integrates differential privacy with SNNs by leveraging gradient noise and discrete spike sequences, aiming to enhance model robustness against privacy attacks~\cite{wang2022dpsnn}. However, the direct impact of different surrogate gradient functions on the privacy characteristics of SNNs remains unexplored, leaving a gap in understanding their potential role in mitigating privacy risks. Similarly, while prior work has examined quantization for ANN privacy, its implications for SNN privacy have not been thoroughly investigated. Our work bridges these gaps by evaluating how quantization and surrogate gradients impact privacy in SNNs, providing insights into their role in enhancing privacy resilience while maintaining model performance.

 % In \hl{federated learning}, Melis et al.~\cite{melis2019exploiting} and Song and Mittal~\cite{song2021systematic} highlighted MIA risks in distributed and generative models, respectively, with LiRA by Ilyas et al.~\cite{carlini2022membership} and RMIA by Zarifzadeh et al.~\cite{zarifzadeh2024low} further advancing MIA efficacy.
\section{Background}
\subsection{Spiking Neural Networks (SNNs)}
\noindent
SNNs are inspired by biological neural activity and represent a fundamental shift from traditional ANNs' continuous outputs. SNNs operate through discrete spikes, occurring only when a neuron’s membrane potential surpasses a specific threshold. This mechanism incorporates time as an additional dimension in information processing, where spike timing patterns encode neural representations~\cite{schuman2022opportunities}. This spike-based mode of communication supports asynchronous data processing and aligns well with neuromorphic hardware designed for event-driven computation, offering enhanced energy efficiency and reduced latency~\cite{schuman2022opportunities}. SNNs require alternative training mechanisms to overcome the non-differentiability of their spike function~\cite{schuman2022opportunities}.

Surrogate gradients provide a workaround for the non-differentiable spike function by substituting a smooth surrogate function in the backward pass. The function approximates the spiking neuron’s activation and is used to compute gradients during backpropagation. Common surrogate gradients include:

\begin{itemize}
    \item \textbf{Fast Sigmoid:} Approximates the gradient using a sharp sigmoid function. It provides precise gradient updates, enhancing feature representation and learning stability.
    
    \item \textbf{Arctangent:} Employs the derivative of the arctangent function as the gradient, offering smoother updates. However, its less aggressive slope makes it less effective in disrupting predictable patterns.

    \item \textbf{Spike Rate Escape:} A gradient model based on a sigmoid-like function with a decay parameter, effectively introducing noise in spike activations. This makes it useful for improving privacy resilience.

    \item \textbf{Triangular:} Uses a linear approximation for gradients. Its weaker gradient strength often leads to instability in learning.

    \item \textbf{Straight Through Estimator (STE):} Utilizes the gradient of a fast sigmoid function during the backward pass, while maintaining a unit derivative for simplicity. This method balances computational efficiency with gradient approximations.
\end{itemize}

Surrogate functions enable gradient flow in SNNs despite the non-differentiability of spike events. They introduce variability through spike timing distributions and gradient approximations, a mechanism aligning with DPSGD, where privacy relies on systematic noise injection during training. So unlike DPSGD’s explicit noise addition, SNNs   inherently generate randomness, which may provide privacy benefits without incurring the performance penalties of DPSGD..

% These surrogate functions enable gradient flow through SNNs, supporting effective learning despite the non differentiability of spike events. The inherent stochasticity and temporal sparsity of SNN activations generate output variability that may serve as a natural privacy preserving mechanism. This characteristic shares fundamental principles with DPSGD, where privacy preservation relies on systematic noise injection during training. Unlike DPSGD's explicit noise addition, SNNs' spike based computations produce natural randomness through both spike timing distributions and surrogate gradient approximations. Such intrinsic variability suggests potential privacy advantages without incurring the performance penalties typically associated with DPSGD implementations. 
\subsection{Membership Inference Attack (MIA)}

% \hl{Your Figure 1 is exactly the same as your Figure 2 in PoPET paper. This is not allowed. you need permission to reuse figure, even your own figure. please redraw the key pieces.}

\begin{figure}[ht!]
    \centering
    \includegraphics[width=1\linewidth, height=0.55\linewidth]{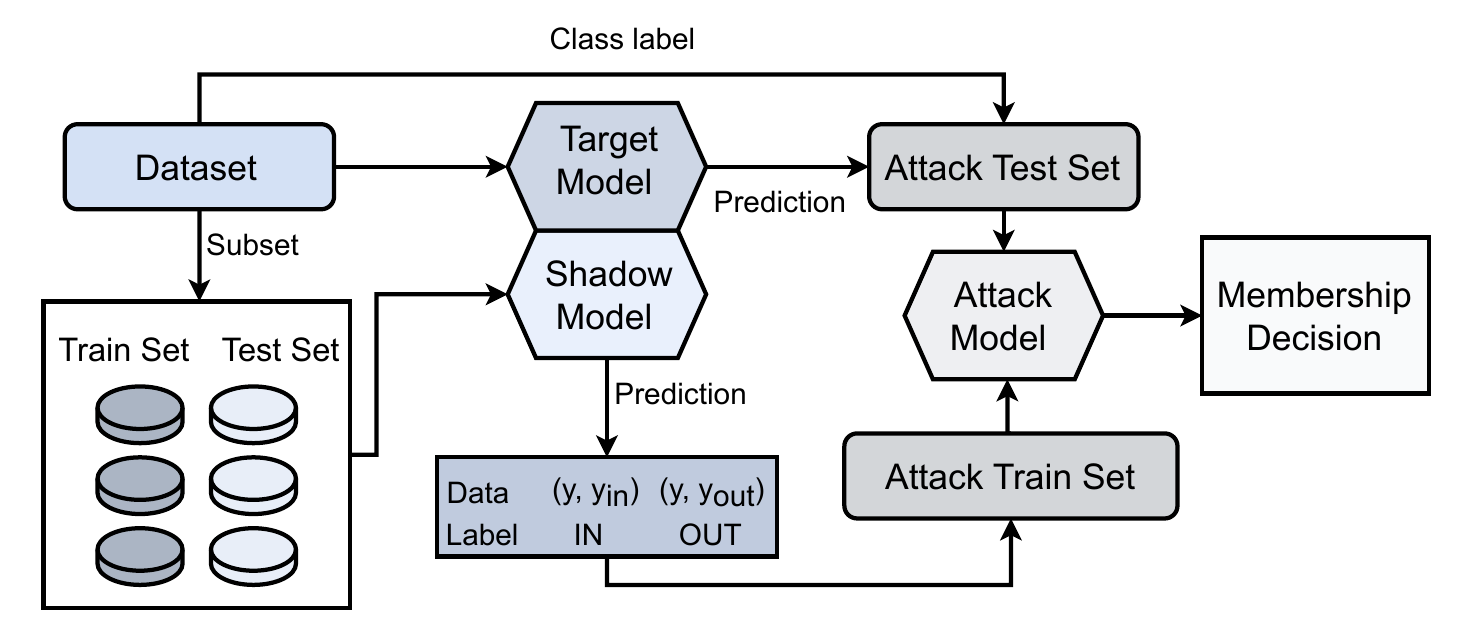}
    \caption{Membership Inference Attack (MIA) Framework}
    \label{fig:mia_24}
\end{figure}
\noindent  
MIA is a privacy attack that enables adversaries to infer the presence of individual data samples in a ML model's training dataset~\cite{shokri}. This attack exploits differences in the model's behavior on training and non-training data.~\cite{rahman2018membership}. Models tend to exhibit higher confidence or different error patterns for samples they have encountered during training, compared to new unseen data~\cite{nguyen2015deep}. By analyzing these patterns, attackers can infer sensitive information about training data, which can lead to privacy breaches~\cite{de2020overview}.

MIA involves training attack models or employing statistical tests to distinguish between the model responses on training and non-training data. The success of these attacks largely depends on the model's overfitting to its training data and the distinctiveness of the model's responses to individual samples. MIAs not only breach data privacy but also expose weaknesses in a model's ability to generalize~\cite{gomm2000case}.

The MIA framework in our experiments involves a two-model approach (Figure~\ref{fig:mia_24}) consisting of a target model and a shadow model, each of the same architecture described in Section~\ref{subsec:model_arch_datasets}.  The process begins with a target model trained on the dataset of interest, which the adversary aims to analyze. To approximate the target model's behavior, a shadow model is trained on an 80\% subset of the same dataset, mimicking the target’s architecture and learning parameters. The attack training dataset is generated using the shadow model, where predictions on its training data are labeled as 'IN' and those on its test data as 'OUT.' The attack test dataset is then constructed from the target model’s predictions, following the same labeling scheme. An SVM with a Radial Basis Function (RBF) kernel serves as the attack model, trained on the constructed attack dataset. For ANNs, logits from the fully connected layer are used as input features, while for SNNs, membrane potentials from the final time step are utilized. To mitigate class imbalance, undersampling is applied during training. The trained attack model is then tested on the target model’s predictions, determining whether a given sample was part of the training set ('IN') or not ('OUT'), quantifying its vulnerability to MIAs.

% \hl{MP: You need to describe the figure. you have this figure in your paper without describing it at all..}

\begin{figure*}[ht]
    \centering
    \includegraphics[width=0.85\linewidth, height=0.50\linewidth]{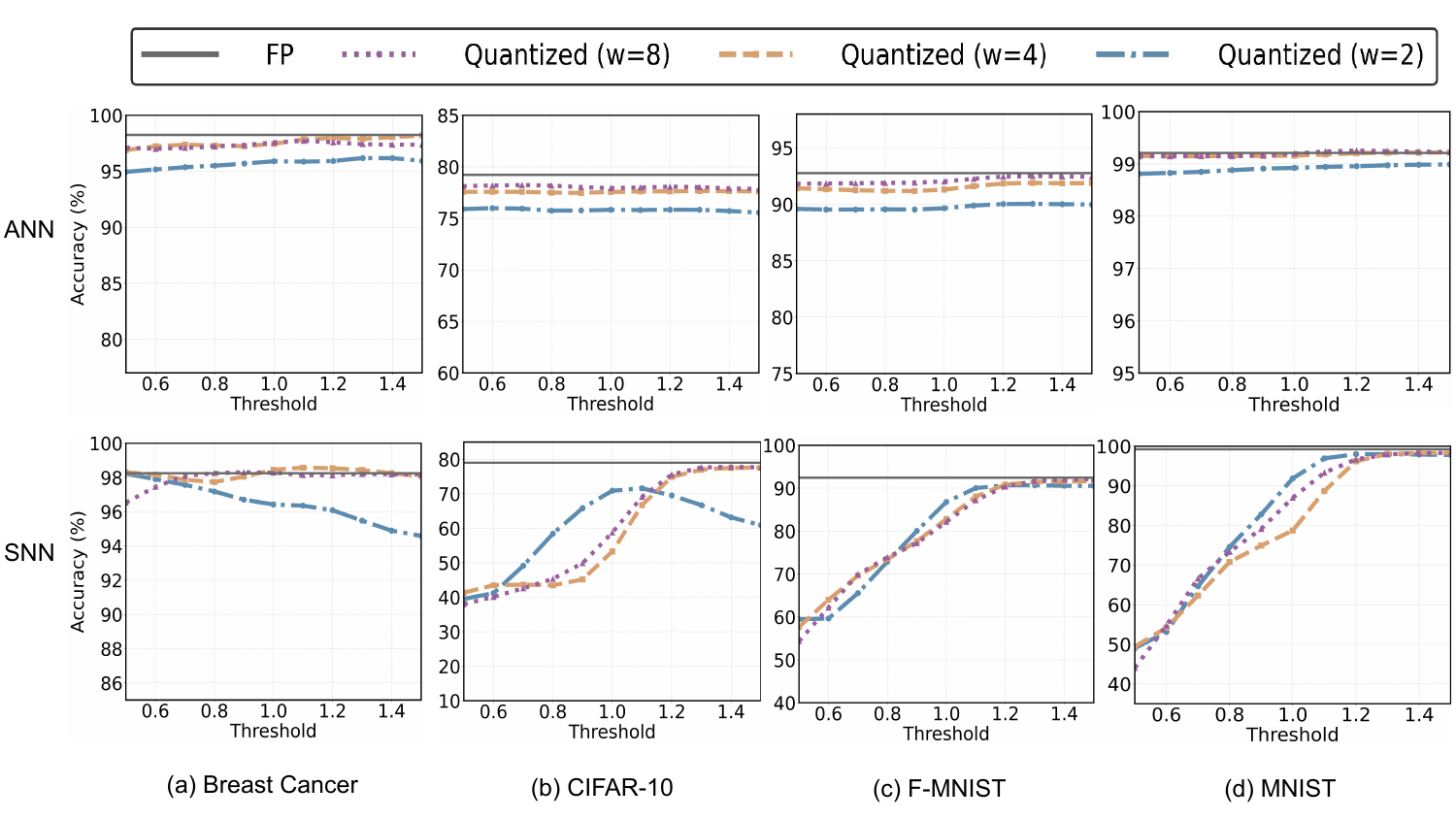}
    \caption{Activation Quantization impact on Model Accuracy on (a) Breast Cancer (b) CIFAR-10, (c) F-MNIST, and (d) MNIST. The grey solid line represents the Full Precision (FP) model, while the purple dotted, orange dashed, and blue dash-dotted lines correspond to the quantized models with bit precisions of w=8, w=4, and w=2 respectively.} 
    \label{fig:act_acc}
\end{figure*}

\begin{figure*}[ht]
    \centering
    \includegraphics[width=0.85\linewidth, height=0.50\linewidth]{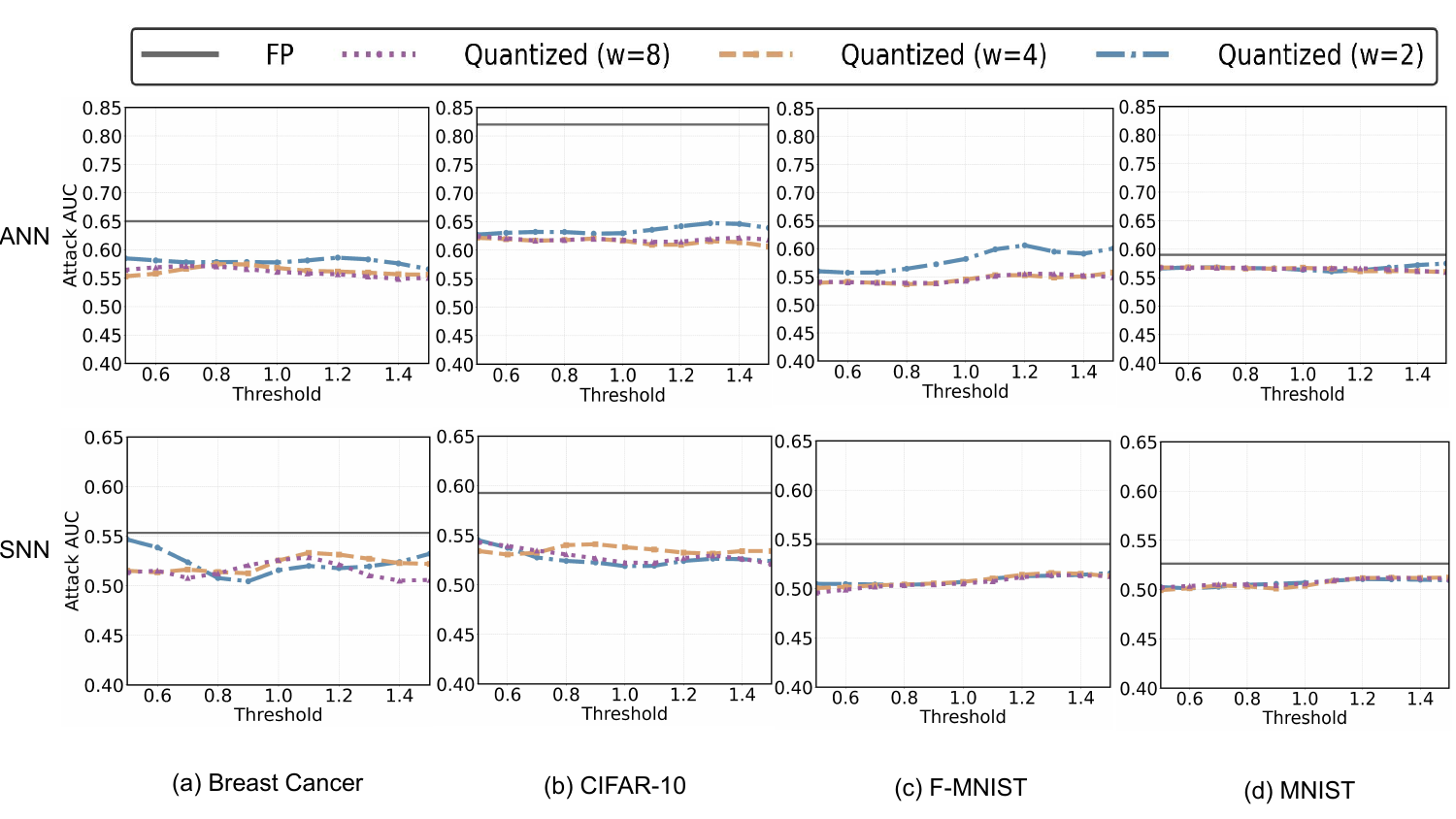}
    \caption{Activation Quantization impact on Privacy Vulnerability  on (a) Breast Cancer (b) CIFAR-10, (c) F-MNIST, and (d) MNIST. The grey solid line represents the Full Precision (FP) model, while the purple dotted, orange dashed, and blue dash-dotted lines correspond to the quantized models with bit precisions of w=8, w=4, and w=2 respectively.}
    \label{fig:act_auc}
\end{figure*}

\subsection{Quantization}
% \subsection{\textbf{Surrogate Gradient}}
\noindent
Quantization reduces the precision of model parameters (weights) and layer activations, typically from 32-bit floating point to lower bit integers~\cite{guo2018survey}. This technique lowers memory usage, accelerates inference, and reduces energy consumption, facilitating the deployment of Deep Neural Networks (DNNs) on resource-constrained devices~\cite{krestinskaya2023towards}. Additionally, the noise inherently introduced by quantization acts as an implicit regularization mechanism, mitigating overfitting and improving generalization~\cite{kang2024effect}.

In this work, we investigate weight quantization and activation quantization and their impact on privacy in ANNs and SNNs.

\textbf{Activation Quantization:} This method reduces the precision of neuron outputs, constraining activations to discrete levels:
\begin{equation}
    a_q = \text{round}\left(\frac{\text{clip}(a, a_{\text{min}}, a_{\text{max}}) - a_{\text{min}}}{\Delta}\right) \cdot \Delta + a_{\text{min}}.
\end{equation}
Here, \( \Delta = \frac{a_{\text{max}} - a_{\text{min}}}{2^k - 1} \) ensures uniform quantization within the activation range \([a_{\text{min}}, a_{\text{max}}]\), where \( k \) represents the bit width of quantization.

\textbf{Weight Quantization:} Weight quantization maps weights \( w \) to a discrete range, optimizing storage and computation:
\begin{equation}
    Q(w) = \Delta \cdot \text{clip}\left(\text{round}\left(\frac{w}{\Delta}\right), 0, 2^k - 1\right),
\end{equation}
where \( \Delta = \frac{w_{\text{max}} - w_{\text{min}}}{2^k - 1} \) defines the step size for a bit-width \( k \), ensuring both positive and negative weights are accounted for.

% \textbf{Weight Quantization:} Weight quantization maps weights \( w \) to a discrete integer range, optimizing storage and computational efficiency. A common formulation for this process is:
% \begin{equation}
%     Q(w) = \Delta \cdot \text{clip}\left(\text{round}\left(\frac{w}{\Delta}\right), 0, 2^k - 1\right),
% \end{equation}
% where \( \Delta = \frac{w_{\text{max}} - w_{\text{min}}}{2^k - 1} \) is the step size, \( w_{\text{max}} \) and \( w_{\text{min}} \) are the range of the weights, and \( k \) is the bit-width. For datasets or models with weights spanning both positive and negative values, the range 
% \([w_{\text{min}}, w_{\text{max}}]\) ensures that negative weights are also accounted for during quantization. This transition reduces precision by representing weights within a constrained range, enabling efficient computation on hardware designed for lower precision arithmetic.
\begin{table}[t]
\footnotesize
\renewcommand{\arraystretch}{1.5}  % Adjust row spacing
\setlength{\tabcolsep}{4pt}  % Adjust column spacing
\begin{center}
\caption{Model Architecture and Configuration}
\label{tab:architectures}
\begin{tabular}{p{1cm}p{1cm}p{1cm}p{4.5cm}}  % Adjusted column widths
\hline
\textbf{Dataset} & \textbf{Network} & \textbf{Model} & \textbf{Structure} \\
\hline
CIFAR10 \newline MNIST \newline FMNIST & ConvNet & ANN & 
2 Conv layers (32, 64 filters) \newline
2 MaxPool layers \newline
2 FC layers (1000, num\_classes) \newline
ReLU activations \\[8pt]
\cline{3-4}
 &  & SNN & 
2 Conv layers (32, 64 filters) \newline
2 MaxPool layers \newline
LIF neurons ($\beta = 0.95$) \newline
Temporal processing ($T = 25$) \\[8pt]

\hline
Iris \newline Breast-Cancer & FCNet & ANN & 
2 FC layers (Input, 1000, num\_classes) \newline
ReLU activations \\[8pt]
\cline{3-4}
 &  & SNN & 
2 FC layers (Input, 1000, num\_classes) \newline
LIF neurons ($\beta = 0.95$) \newline
Temporal processing ($T = 25$) \\
\hline
\end{tabular}
\end{center}
\end{table}

% \begin{table}[t]
% \footnotesize
% \setlength{\tabcolsep}{4pt}  % Adjust column spacing
% \begin{center}
% \caption{Model Architectures and Configurations}
% \label{tab:architectures}
% \begin{tabular}{p{1
% cm}p{0.9cm}p{4.5cm}p{1.1cm}}  % Adjusted column widths
% \hline
% Network & Variant & Structure & Params* \\
% \hline
% \multicolumn{4}{c}{} \\
% ConvNet & ANN & $\bullet$ 2 Conv layers (32, 64 filters) \newline
%        $\bullet$ 2 MaxPool layers \newline
%        $\bullet$ 2 FC layers (1000, num\_classes) \newline
%        $\bullet$ ReLU activations & $\sim$2.3M \\[8pt]

% \hline
% \multicolumn{4}{c}{} \\
%  & SNN & $\bullet$ 2 Conv layers (32, 64 filters) \newline
%        $\bullet$ 2 MaxPool layers \newline
%        $\bullet$ LIF neurons ($\beta = 0.95$) \newline
%        $\bullet$ Temporal processing ($T = 25$) & $\sim$2.3M \\[8pt]
% \hline
% \multicolumn{4}{c}{} \\
% FCNet & ANN & 
%        $\bullet$ 2 FC layers (Input, 1000, num\_classes) \newline
%        $\bullet$ ReLU activations & 8K–33K \\[8pt]
% \hline
% \multicolumn{4}{c}{} \\
%  & SNN & 
%        $\bullet$ 2 FC layers (Input, 1000, num\_classes) \newline
%        $\bullet$ LIF neurons ($\beta = 0.95$) \newline
%        $\bullet$ Temporal processing ($T = 25$) & 8K–33K \\
% \hline
% \multicolumn{4}{c}{} \\
% \end{tabular}
% {\small
% \begin{flushleft}
% *Params vary with input channels (1/3) for ConvNet or input features for FCNet.
% \end{flushleft}}
% \end{center}
% \end{table}

\begin{table*}[htbp]
\centering
\resizebox{\textwidth}{!}{%
\begin{tabular}{l l c c c c c c c}
\toprule
\textbf{Dataset} & \textbf{Method} & \multicolumn{3}{c}{\textbf{ANN}} & \multicolumn{3}{c}{\textbf{SNN}} \\
\cmidrule(lr){3-5} \cmidrule(lr){6-8}
 &  & \textbf{Train Accuracy} & \textbf{Test Accuracy} & \textbf{MIA AUC} & \textbf{Train Accuracy} & \textbf{Test Accuracy} & \textbf{MIA AUC} \\
\midrule
\textbf{MNIST} & Full precision & 99.96(±0.03)\% & 99.21(±0.01)\% & 0.5900(±0.008) & 99.95(±0.05)\% & 99.22(±0.02)\% & 0.5264(±0.011) \\
 & quantized(w=8) & 99.88(±0.05)\% & 99.16(±0.02)\% & 0.5674(±0.007) & 99.93(±0.03)\% & 99.16(±0.03)\% & 0.5101(±0.012) \\
 & quantized(w=4) & 99.85±(0.04)\% & 99.12(±0.04)\% & 0.5800(±0.003) & 99.91(±0.02)\% & 99.21(±0.05)\% & 0.5117±(0.009) \\
 & quantized(w=2) & 99.57(±0.02)\% & 99.07(±0.03)\% & 0.5667(±0.009) & 99.88±(0.03)\% & 99.11(±0.04)\% & 0.4993(±0.005) \\
\midrule
\textbf{FMNIST} & Full precision & 99.52(±0.13)\% & 92.77(±0.20)\% & 0.6400(±0.011) & 99.42(±0.32)\% & 92.44(±0.19)\% & 0.5453(±0.008) \\
 & quantized(w=8) & 95.98(±0.08)\% & 92.30(±0.10)\% & 0.5397(±0.010) & 96.49(±0.91)\% & 92.00(±0.10)\% & 0.5022(±0.003) \\
 & quantized(w=4) & 95.47(±0.72)\% & 92.15(±0.09)\% & 0.5406(±0.005) & 96.44(±0.53)\% & 91.97(±0.17)\% & 0.4970(±0.010) \\
 & quantized(w=2) & 92.61(±0.20)\% & 90.69(±0.18)\% & 0.5295(±0.004) & 93.37(±0.51)\% & 91.36(±0.18)\% & 0.5024(±0.005) \\
\midrule
\textbf{CIFAR10} & Full precision & 99.24(±0.08)\% & 79.20(±0.43)\% & 0.8200(±0.095) & 99.13(±0.45)\% & 78.99(±0.33)\% & 0.5927(±0.005) \\
 & quantized(w=8) & 79.61(±0.20)\% & 77.78(±0.28)\% & 0.6800(±0.015) & 72.37(±0.44).\% & 73.39(±0.13)\% & 0.5250(±0.022) \\
 & quantized(w=4) & 79.56(±0.41)\% & 77.71(±0.39)\% & 0.7000(±0.021) & 72.54(±0.29)\% & 72.92(±0.51)\% & 0.5405(±0.011) \\
 & quantized(w=2) & 71.35(±0.24)\% & 73.37(±0.15)\% & 0.6200(±0.015) & 66.41(±1.09)\% & 68.59(±0.55)\% & 0.5280(±0.002) \\
\midrule
\textbf{Iris} & Full precision & 98.33(±2.34)\% & 100.0(±0.00)\% & 0.7700(±0.13) & 100.0(±0.00)\% & 100.0(±0.00)\% & 0.5728(±0.020) \\
 & quantized(w=8) & 82.78(±2.02)\% & 93.33(±2.94)\% & 0.6500(±0.070) & 79.44(±7.33)\% & 95.56(±1.92)\% & 0.5234(±0.012) \\
 & quantized(w=4) & 85.56(±1.05)\% & 91.11(±1.92)\% & 0.7163(±0.160) & 84.17(±2.47)\% & 94.44(±3.06)\% & 0.5304(±0.045) \\
 & quantized(w=2) & 84.06(±2.20)\% & 91.11(±3.06)\% & 0.6400(±0.013) & 81.94(±4.18)\% & 92.22(±6.71)\% & 0.5380(±0.020) \\
\midrule
\textbf{Breast Cancer} & Full precision & 99.34(±0.04)\% & 98.25(±0.00)\% & 0.6500(±0.008) & 100.0(±0.00)\% & 98.25(±0.24)\% & 0.5534(±0.017) \\
 & quantized(w=8) & 97.51(±0.13)\% & 97.96(±0.04)\% & 0.5500(±0.006) & 99.09(±0.64)\% & 
 98.25(±0.00)\% & 0.4983(±0.003) \\
 & quantized(w=4) & 97.58(±0.57)\% & 97.96(±0.49)\% & 0.5404(±0.010) & 97.69(±0.36)\% & 97.66(±0.51)\% & 0.5099(±0.021) \\
 & quantized(w=2) & 97.80(±0.46)\% & 97.54(±0.40)\% & 0.5800(±0.013) & 97.87(±0.48)\% & 
 97.37(±0.00)\% & 0.5088(±0.021) \\
\bottomrule
\end{tabular}%
}
\caption{Weight Quantization Impact on Privacy Vulnerability and Model Performance: Comparison between ANN and SNN models across different datasets.}
\label{table:weight_quan}
\end{table*}

Quantized models require gradient approximation during training. In ANNs, Quantization-Aware Training (QAT) updates quantized parameters while maintaining gradient flow. In SNNs, non-differentiable spike functions necessitate surrogate gradients approximations. This study applies QAT  with \textit{fast sigmoid} to examine their impact on privacy vulnerability while maintaining accuracy.

\section{Experimental Setup}

% \hl{MP: I'm not sure about calling this section, Research methods. There is not much details on the method or theroy. Maybe you can rename it to "Experimental Setup"}

% This section describes the experimental framework, including model architectures, datasets, quantization methods, and surrogate gradient configurations. All implementations utilized PyTorch~\cite{ketkar2021introduction} and SNNtorch~\cite{snntorch}.

\subsection{Model Architectures and Datasets}\label{subsec:model_arch_datasets}

\noindent
This study evaluates CIFAR10~\cite{cifar10}, MNIST~\cite{lecun2010mnist}, FMNIST~\cite{xiao2017fashion}, Iris~\cite{omelina2021survey}, and Breast Cancer~\cite{misc_breast_cancer_14}, implemented using PyTorch~\cite{ketkar2021introduction} and SNNtorch~\cite{eshraghian2021training}. For CIFAR10, MNIST, and FMNIST we follow their standard splits (50,000 training, 10,000 testing), while for Iris and Breast Cancer we apply an 80-20 stratified split.

We preprocess CIFAR10 by normalizing the data to a mean of [0.5, 0.5, 0.5], applying random cropping with a 4-pixel padding and 50\% horizontal flipping. For MNIST and FMNIST, we resize the images to 28×28 pixels, convert them to grayscale, and normalize them with a mean of 0 and standard deviation of 1, without augmentations. For Iris and Breast Cancer datasets, we standardize the features using \texttt{StandardScaler}~\cite{scikit-learn}.

We use Convolutional Neural Networks (CNNs) (ConvNet) for CIFAR10, MNIST, and FMNIST , and Fully Connected Networks (FCNets) for Iris and Breast Cancer. We summarize the detailed configurations for both ANNs and SNNs in Table~\ref{tab:architectures}. For activation functions, we use ReLU for ANNs and Leaky Integrate and Fire (LIF) neurons (\( \beta = 0.95\)) for SNNs, which operate over a time step of 25 to propagate spikes.

\subsection{Quantization Methods}

% \hl{MP: Why don't you clearly start paragraphs one with weight quantization and the other with activation? to follow the same structure as your previous section? }

% \hl{MP: In this section you have described first weight quantization and then activation quantization. In the result and figures you first have activation quantization. Please be consistent}

% \hl{MP: What is your gradient of choice in SNNs when you do activation quantization here? In ANNs you had QAT. In SNNs you said you applied QAT with different surrogate gradients. in Fig. 3 where you have activation quantization what surr. gradient did you use for SNNs?}

% \hl{MP: In figure, it is really hard to see the lines. trends are clear both colors and dashed lines are too vague. For full precision you can show it with solid line and you need to also write it in the caption.}

% \hl{MP: Lines in Fig 4 are solid, in Fig 3 are dashed. I prefer Fig 4 to 3. Please be as consistent as possible and make them all solid. }

% \hl{MP: In figure 4 it is hard to see FP line. Please make the FP line a bit wider in both figures}

% \hl{MP: What is w in activation quantization figures 2 and 3?? :D w is weight? shouldn't that be for weight quantization (figure 4 only)?}

% \hl{MP: What is FRR and TRR in Figyre 4? The axes titles are too small. you need to increase font size for the axes numbers and titles}

\noindent
We implement activation quantization differently for ANNs and SNNs. In ANNs, we quantize activations using \texttt{brevitas.nn} QuantReLU~\cite{brevitas}, while in SNNs, we develop a custom \texttt{state\_quant} function within snnTorch~\cite{eshraghian2021training} to quantize membrane potentials. To assess the effect of activation precision, we apply thresholds ranging from 0.5 to 1.5 before quantization, clipping activations and setting an upper bound on magnitudes included in the quantized representation. Lower thresholds impose stricter clipping, reducing the range of preserved activations and potentially limiting representational capacity. In contrast, higher thresholds retain a broader range of activations, which may enhance expressiveness but could also increase vulnerability to privacy attacks.

We apply weight quantization using \texttt{brevitas.nn} for both ANNs and SNNs, reducing parameter precision to 2-bit, 4-bit, and 8-bit levels. This process discretizes model weights, minimizing storage requirements and computational cost while potentially influencing privacy vulnerability.
% Two quantization techniques were applied: weight quantization using \texttt{brevitas.nn}~\cite{brevitas} for both ANNs and SNNs, and activation quantization, implemented differently for each architecture. For ANNs, activations were quantized using \texttt{brevitas.nn} QuantReLU, while for SNNs, a custom \texttt{state\_quant} function was developed within snnTorch~\cite{snntorch} \hl{to quantize membrane potentials.}
% Experiments were conducted at 2-bit, 4-bit, and 8-bit precision to analyze how reduced numerical representation affects both model accuracy and privacy vulnerability. For Activation Quantization, thresholds ranging from 0.5 to 1.5 were applied to clip activation values before quantization, setting the upper bound on activation magnitudes included in the quantized representation. Lower thresholds impose stricter clipping, reducing the range of preserved activations and potentially limiting representational capacity. In contrast, higher thresholds retain a broader range of activations, which may enhance expressiveness but could also increase vulnerability to privacy attacks.

\subsection{Surrogate Gradient Configurations}

\noindent
We use five surrogate gradients in our experiments: \textit{Fast Sigmoid}, \textit{Straight Through Estimator}, \textit{Arctangent}, \textit{Spike Rate Escape}, and \textit{Triangular}. Among these, we specifically employ \textit{Fast Sigmoid} for quantization studies~\cite{eshraghian2021training}. We set the \textit{Fast Sigmoid} gradient slope to 25, while  \textit{Arctangent (atan)}  uses an alpha value of 2 which controls the steepness of its curve. For \textit{Spike Rate Escape}, we apply a beta parameter of 1 which regulates the escape rate with a slope of 25. We use \textit{Triangular} and \textit{Straight Through Estimator (STE)} with their default configurations in the framework.
\noindent

\section{Results \& Discussion}
\noindent
This section presents the evaluation of how activation and weight quantization as well as surrogate gradients impacts SNN privacy, focusing on MIA vulnerability and performance trade-offs in comparison to ANNs. Privacy vulnerability is measured via ROC-AUC, while accuracy captures performance impact.

% \hl{general comment: Quantiziation in ANNs started due to the need for energy efficiency. The fact that FP SNN is still more privacy preserving that quantizied ANN as standalone might not have enough merit if not compared with energy savings of SNNs vs quantizied SNNs and quantizied ANNs. You need add a note on this or even add a column comparing the energy numbers from [34]}

\subsection{Activation Quantization}

\noindent
% Figures~\ref{fig:act_acc} and~\ref{fig:act_auc} illustrate the impact of activation quantization on model accuracy and attack AUC across datasets.

\noindent
\textit{Model Performance:} 
Accuracy degradation due to activation quantization is observed in both ANNs and SNNs, but the overall drop from full precision to quantized models with 8-bits, 4-bits and 2-bits remains relatively small (Figure~\ref{fig:act_acc}). In ANNs, accuracy decreases more noticeably at lower bit-widths, particularly at 2-bit, where reduced precision compresses activation ranges, limiting representational capacity and reducing the distinction between logits. However, across each bit-width, accuracy remains relatively stable across different threshold levels, indicating that thresholding has minimal impact on model performance. This is because  ReLU activations, operate in a continuous space where moderate clipping does not significantly alter feature representations, allowing the network to maintain performance despite threshold variations. 
% For example, on CIFAR-10~\cite{cifar10}, accuracy dropped from 79.20\% (full precision) to 75\% in 2-bit models. This decline reflects the loss of precision in model parameters, compressing activation ranges and reducing the distinction between logits. 

In SNNs, accuracy trends vary across datasets, as shown in Figure~\ref{fig:act_acc}. For 4-bit and 8-bit quantized models, higher thresholds consistently improve accuracy across all datasets by retaining a broader activation range, preserving useful signal information. The stochastic nature of SNNs further mitigates quantization noise, ensuring stable performance regardless of dataset complexity. However, 2-bit quantized models show different behaviors. In Breast Cancer, accuracy starts near full precision at lower thresholds but degrades as thresholds increase due to amplified noise overwhelming simpler patterns. In CIFAR-10, accuracy initially improves with thresholding but drops at higher thresholds as excessive noise disrupts feature representation, reducing classification performance.

% While full-precision SNNs achieved 78.99\% on CIFAR-10 and 98.25\% on Breast Cancer, lower-bit models exhibited non-monotonic behavior. \hl{In CIFAR-10, the 2-bit model reached 68\% accuracy at a 0.6 threshold, outperforming higher bit-width models but dropping to 60\% at higher thresholds. Similarly, in Breast Cancer, 2-bit models maintained 98\% accuracy at low thresholds but dropped to 94\% at higher thresholds, with 4-bit and 8-bit models stabilizing at 96–97\%.}  \hl{This suggests that while SNNs initially handle quantization noise well, higher thresholds exacerbate accuracy degradation due to cumulative spike dynamics.}

\textit{MIA Vulnerability: }Quantization reduces attack vulnerability in ANNs compared to full precision as seen in Figure~\ref{fig:act_auc}. Quantized models with 2-bit, 4-bit, and 8-bit precision all exhibit lower vulnerability than the full-precision model. However, among them, 2-bit quantized model is the most vulnerable across all datasets, while  4-bit and 8-bit quantized model provide better privacy protection.  This occurs because extreme quantization (2-bit precision) severely limits representational capacity, making activations more uniform and easier to infer in ANNs, thereby increasing susceptibility to attacks. 

% \hl{On CIFAR-10, attack AUC dropped from 0.82 (full precision) to 0.62 at 8-bit and 4-bit precision (24.4\% reduction). However, at 2-bit precision, AUC slightly increased to 0.65}, likely due to distortions that inadvertently retain exploitable characteristics.

In SNNs, activation quantization minimizes attack vulnerability compared to full precision across all datasets as well. However, unlike in ANNs,  2-bit, 4-bit, and 8-bit quantized SNN models showed  somewhat similar privacy vulnerability, specially in F-MNSIT and MNIST datasets. This suggests that the stochastic nature of spike-based encoding inherently limits the granularity of information available to attackers, irrespective of precision levels. The inherent randomness in spike timing and event-driven processing disrupts predictable activation patterns, making the additional noise introduced by quantization less impactful in distinguishing training from non-training data. This means that while quantization is still effective in lowering attack success rates, further reducing bit width does not provide additional privacy benefits for these datasets.

\begin{table*}[htbp]
\centering
\scriptsize % Reduce font size
\renewcommand{\arraystretch}{0.9} % Reduce row height
\setlength{\tabcolsep}{4pt} % Reduce column spacing
\begin{tabular}{p{2cm} p{4cm} p{1.5cm} p{1.5cm} p{1.5cm}}
\toprule
\textbf{Dataset} & \textbf{Surrogate Gradients} & \multicolumn{3}{c}{\textbf{SNN}} \\
\cmidrule(lr){3-5}
 &  & \textbf{Train Accuracy} & \textbf{Test Accuracy} & \textbf{MIA AUC} \\
\midrule
\textbf{MNIST} & Fast Sigmoid(slope,k=25) & 99.88(±0.01)\% & 99.22(±0.04)\% & 0.518(±0.001) \\
 & aTan(alpha=2) & 99.96(±0.02)\% & 99.25(±0.06)\% &  0.547(±0.006)  \\
 & Spike Rate Escape(beta=1, slope=25) & 99.97(±0.01)\% & 99.25(±0.03)\% & 0.508(±0.008) \\
 & Triangular & 75.16(±0.09)\% & 76.34(±0.14)\% & 0.503(±0.003) \\
 & Straight Through Estimator & 99.57(±0.04)\% & 98.79(±0.07)\% & 0.528(±0.005) \\
\midrule
\textbf{FMNIST} & Fast Sigmoid(slope,k=25) & 99.45(±0.04)\% & 91.97(±0.12)\% & 0.518(±0.008) \\
 & aTan(alpha=2) & 99.58(±0.03)\% & 92.18(±0.14)\% & 0.547(±0.002) \\
 & Spike Rate Escape(beta=1, slope=25) & 99.74(±0.09)\% & 92.23(±0.11)\% & 0.523(±0.016) \\
 & Triangular & 78.79(±0.24)\% & 79.49(±0.20)\% & 0.498(±0.011) \\
 & Straight Through Estimator & 91.55(±0.13)\% & 89.69(±0.21)\% & 0.512(±0.003) \\
\midrule
\textbf{CIFAR10} & Fast Sigmoid(slope,k=25) & 82.90(±0.46)\% & 78.04(±0.40)\% & 0.535(±0.010) \\
 & aTan(alpha=2) & 87.56(±0.41)\% & 78.99(±0.38)\% & 0.561(±0.010) \\
 & Spike Rate Escape(beta=1, slope=25) & 89.92(±0.39)\% & 79.95(±0.33)\% & 0.567(±0.013) \\
 & Triangular & 21.43(±0.76)\% & 23.92(±0.86)\% & 0.550(±0.036) \\
 & Straight Through Estimator & 53.10(±0.66)\% & 62.49(±0.54)\% & 0.644(±0.034) \\
\midrule
\textbf{Iris} & Fast Sigmoid(slope,k=25) & 82.50(±2.46)\% & 90.00(±6.34)\% & 0.654(±0.022) \\ 
 & aTan(alpha=2) & 96.67(±1.46)\% & 93.33(±0.50)\% & 0.563(±0.095) \\
 & Spike Rate Escape(beta=1, slope=25) & 96.67(±1.00)\% & 100.00(±0.00)\% & 0.543(±0.036) \\
 & Triangular & 100.00(±0.00)\% & 100.00(±0.00)\% & 0.510(±0.029) \\
 & Straight Through Estimator & 97.50(±0.04)\% & 100.00(±0.00)\% & 0.542(±0.108) \\
\midrule
\textbf{Breast Cancer} & Fast Sigmoid(slope,k=25) & 100.00(±0.00)\% & 100.00(±0.00)\% & 0.494(±0.013) \\
 & aTan(alpha=2) & 100.00(±0.00)\% & 97.37(±0.11)\% & 0.538(±0.015) \\
 & Spike Rate Escape(beta=1, slope=25) & 100.00(±0.00)\% & 97.37(±0.11)\% & 0.512(±0.026) \\
 & Triangular & 97.58(±0.12)\% & 98.25(±0.00)\% & 0.497(±0.022) \\
 & Straight Through Estimator & 99.78(±0.12)\% & 98.25(±0.00)\% & 0.482(±0.024) \\
\bottomrule
\end{tabular}
\caption{Impact of training SNNs with different surrogate gradients on Privacy Vulnerability and Model Performance across different datasets.}
\label{table:sg}
\end{table*}

\subsection{Weight Quantization}
\noindent
% Table~\ref{table:weight_quan} and Figure~\ref{fig:weight_auc} illustrate the effects of weight quantization on model performance and MIA vulnerability across datasets.

\noindent
\textit{Model Performance:} Weight quantization leads to accuracy degradation in both ANNs and SNNs  from full precision models  as shown in Table~\ref{table:weight_quan}. In ANNs, the most pronounced drop is observed in 2-bit quantized models, similar to the impact of activation quantization. Extreme quantization severely limits weight precision, resulting in coarser updates that reduce representational capacity and hinders the model’s ability to learn fine-grained patterns. This effect is evident in CIFAR-10, where accuracy declines more sharply at lower bit width of 2. The higher complexity of CIFAR-10 requires finer weight precision to capture intricate features, and 2-bit quantization struggles to retain the necessary detail for accurate classification. 

In SNNs, weight quantization similarly affects accuracy  as shown in Table~\ref{table:weight_quan}. The most pronounced drop is observed in 2-bit quantized models, where  limited weight precision reduces the model's capacity to represent detailed spike-based patterns. Like ANNs, this is especially evident in CIFAR-10 where, 2-bit quantized model struggles to encode intricate features effectively, resulting in a sharper accuracy decline compared to 4-bit and 8-bit models.

% SNNs showed comparable sensitivity to weight quantization. On CIFAR-10, accuracy decreased \hl{from 78.99\% (full precision) to 73.39\% at 8-bit and 68.59\% at 2-bit}, with similar trends observed across datasets. \hl{Lower bit-widths introduced quantization noise, impairing representational capacity and generalization}.

\textit{MIA Vulnerability:} Weight quantization consistently reduces MIA vulnerability in ANNs compared to full-precision models across all datasets (Table~\ref{table:weight_quan}). In most cases, models quantized to 2-bit, 4-bit, and 8-bit levels demonstrate comparable AUC values, effectively obscuring training data from attackers through noise introduced by quantization. Notably, 2-bit quantized models exhibit the lowest attack AUC values across datasets, representing a trend opposite to that observed with activation quantization. This difference arises from how different quantization noise affects model representations. For weight quantization, 2-bit precision disrupts parameter-level patterns, introducing randomness that hampers an attacker’s ability to infer sensitive training data. By contrast, activation quantization with 2-bit precision compresses activation ranges excessively, resulting in uniform outputs that are easier to predict, thereby increasing vulnerability.

% , with lower bit-widths yielding greater AUC reductions. On FMNIST, AUC decreased from 0.64 (full precision) to 0.56 at 8-bit, 0.54 at 4-bit, and 0.53 at 2-bit, marking reductions of 12.5\%, 15.6\%, and 17.2\%, respectively. Similar trends were observed across MNIST, CIFAR-10, and Breast Cancer, demonstrating weight quantization’s role in limiting model output granularity and reducing privacy risks.

In SNNs, weight quantization reliably lowers MIA vulnerability across all datasets compared to full-precision models  though reductions are smaller due to the already lower baseline AUC. Models quantized to 2-bit, 4-bit, and 8-bit levels demonstrate similar AUC values which indicates that weight quantization has a uniformly beneficial impact on reducing vulnerability across all levels of quantization. This trend is similar to the impact observed in activation quantization, where different bit widths also produced closely aligned AUC values. The underlying reason for this consistency lies in the stochastic nature of spike-based computation, which inherently disrupts predictable activation patterns. In SNNs, weight quantization further amplifies this effect by introducing additional noise to synaptic weights, but due to the already high variability in spike timing and membrane dynamics, the additional perturbations from quantization do not significantly alter the model’s susceptibility to MIAs.

From both accuracy and vulnerability perspectives, extreme activation quantization ( 2-bit precision) provides no tangible benefit in neither ANNs nor SNNs. Thus further reducing activation bit width beyond moderate levels (4-bit or 8-bit) is ineffective and unnecessary. 
When comparing the impact of activation and weight quantization in MIA vulnerability for ANNs and SNNs, it is evident that SNNs consistently demonstrate lower MIA vulnerability compared to their ANN counterparts, regardless of quantization levels. Remarkably, even fully precise SNN models, which theoretically should be more vulnerable than quantized models, exhibit lower privacy vulnerability than quantized ANNs at any bit width. This highlights the inherent privacy advantages of SNNs due to their stochastic spike-based encoding and temporal dynamics, which naturally disrupt predictable patterns that attackers exploit.

\subsection{Surrogate Gradients}
\noindent

Surrogate gradients are evaluated from three angles: privacy vulnerability, performance, and trade-off between the two as depicted in Table~\ref{table:sg}.

From the perspective of vulnerability, Spike Rate Escape stands out as the most resilient surrogate gradient across most datasets, effectively lowering attack success rates. Its decay parameter introduces sufficient noise in spike activation patterns, making it harder for attackers to infer training data. In contrast, arctangent often demonstrates the highest vulnerability, likely due to its smoother gradient approximations that fail to sufficiently disrupt predictable patterns. Straight Through Estimator (STE) also exhibits high vulnerability in complex datasets, such as CIFAR-10, where its simplistic gradient approximation inadequately masks sensitive patterns.

In terms of performance, Fast Sigmoid consistently delivers the best results across datasets due to its sharp gradient slopes, allowing precise updates and better feature representation. In contrast, Triangular struggles significantly, performing poorly across datasets. Its weaker gradient approximations may provide insufficient feedback for  parameter updates, especially in feature-rich datasets. Additionally, its inability to leverage the stochastic nature of SNNs effectively may contribute to both lower accuracy and a failure to disrupt predictable patterns. 

When balancing vulnerability and performance, Spike Rate Escape offers the best trade-off, combining strong accuracy with consistently lower vulnerability. Its ability to integrate noise effectively complements its robust feature learning capabilities. In contrast, arctangent and STE fail to strike this balance, as their high vulnerability undermines their moderate performance, making them less effective for scenarios prioritizing both privacy and accuracy.

\section{Conclusion}
\noindent
 The growing implementation of machine learning across sectors like healthcare, finance, and education has raised concerns about potential privacy breaches through inference attacks, particularly when models process sensitive data. Previous research examining privacy characteristics of neural networks has shown that SNNs exhibit better resilience against MIAs compared to traditional ANNs. This study explores how the privacy-preserving characteristics of SNNs can be further enhanced through quantization and by leveraging various surrogate gradient training methods. Our results show that while quantization reduces MIA vulnerability in both SNNs and ANNs,  the privacy advantage of SNNs remains inherent. Notably, even full-precision SNNs exhibit lower vulnerability than quantized ANNs at any bit width, reinforcing the fundamental privacy resilience of spike-based computation over traditional neural architectures. The training of SNNs with different surrogate gradients further highlights their impact on balancing accuracy and privacy. Spike Rate Escape provides the best privacy protection while maintaining strong performance, whereas Arctangent and Straight Through
Estimator (STE) exhibit higher vulnerability. 
Looking forward, we will investigate the energy efficiency of full-precision SNNs compared to quantized ANNs, as the privacy advantage of SNNs holds greater significance when evaluated alongside their energy savings. Additionally, future work will integrate Differentially Private Stochastic Gradient Descent (DPSGD) with quantized ANN and SNN models to explore the synergistic impact of these techniques on both model performance and privacy.

 % This study investigates how quantization affects privacy characteristics in both architectures while examining the impact of different surrogate gradients in SNNs through experiments on CIFAR-10, MNIST, FMNIST and Breast Cancer datasets. Our analysis shows that ANNs exhibit a clear trade off between quantization and privacy: lower precision improves privacy but reduces accuracy. SNNs displayed varied behavior: weight quantization followed similar trends to ANNs, while activation quantization improved privacy compared to full precision but lacked consistent patterns across bit widths. But interestingly, even full precision SNNs provided better privacy protection than quantized ANNs. Among surrogate gradients, spike rate escape achieved optimal balance between accuracy and privacy, while aTan increased vulnerability towards MIA. These results advance our understanding of how architectural choices and training methodologies in neuromorphic systems influence privacy characteristics. Future work will focus on integrating DPSGD with quantized ANN and SNN models to examine the combined impact of quantization and differential privacy on model performance and privacy.
 % \textcolor{blue}{Shay, can you help with the concluding line?}

%Bibliography
\bibliographystyle{IEEEtran}  
% \bibliography{references}  

\end{document}